%% file: main.tex
\documentclass[10pt,twocolumn,letterpaper]{article}

\usepackage[pagenumbers]{cvpr} 

\input{preamble}
\definecolor{cvprblue}{rgb}{0.21,0.49,0.74}
\usepackage[pagebackref,breaklinks,colorlinks,allcolors=cvprblue]{hyperref}

\usepackage{hyperref}
\usepackage{url}
\usepackage{algorithm}
\usepackage{algpseudocode}
\usepackage{graphicx} 
\usepackage{booktabs}
\usepackage{multirow}
\usepackage{wrapfig}
\graphicspath{{figs/}}
\hypersetup{
    colorlinks=false,
    pdfborder={0 0 0}
}

\def\paperID{5875} 
\def\confName{CVPR}
\def\confYear{2026}

\title{Self-Diffusion Driven Blind Imaging}

\author{Yanlong Yang\\
{\tt\small yanlong.yang@my.jcu.edu.au}
\and
Guanxiong Luo\thanks{Equally contributed to this work.}\\
{\tt\small luoguan5@gmail.com}
}

\begin{document}
\maketitle
\input{sec/0_abstract}    
\input{sec/1_intro}
\input{sec/2_Related_Work}
\input{sec/3_method}
\input{sec/4_Exps}
\input{sec/5_Evaluation}

\input{sec/6_Discussion}
{
    \small
    \bibliographystyle{ieeenat_fullname}
    \bibliography{main}
}


\end{document}

%% file: sec/0_abstract.tex
\begin{abstract}
Optical imaging systems are inherently imperfect due to diffraction limits, lens manufacturing tolerances, assembly misalignment, and other physical constraints. In addition, unavoidable camera shake and object motion further introduce non-ideal degradations during acquisition. These aberrations and motion-induced variations are typically unknown, difficult to measure, and costly to model or calibrate in practice. Blind inverse problems offer a promising direction by jointly estimating both the latent image and the unknown degradation kernel. However, existing approaches often suffer from convergence instability, limited prior expressiveness, and sensitivity to hyperparameters.
Inspired by recent advances in self-diffusion, we propose DeblurSDI, a zero-shot, self-supervised blind imaging framework that requires no pre-training. DeblurSDI formulates blind image recovery as an iterative reverse self-diffusion process that begins from pure noise and progressively refines both the sharp image and the blur kernel. Extensive experiments on combined optical aberrations and motion blur demonstrate that DeblurSDI consistently outperforms other methods by a substantial margin.
\end{abstract}

%% file: sec/1_intro.tex
\section{Introduction}
\label{sec:intro}

Optical imaging systems are widely used in various sensors such as cameras \cite{Hasinoff2016,Chen2017HDR}, microscopes \cite{Shaw1991}, telescopes \cite{PlattShack2001}, and LiDARs that operate across wavelengths from visible to infrared. The lens system and the focal-plane image sensor jointly constitute a complete imaging module. In a paraxial, shift-invariant optical system, the captured image can be modeled as 
\begin{equation}
    I(x, y)=O(x,y)\circledast h(x,y),
\end{equation}
where $O(x,y)$ denotes the object’s intensity distribution and $h(x,y)$ is the lens's point spread function (PSF) given by 
\begin{equation}
    h(x,y)=\left|\mathcal{F}\{P(\rho,\theta)\,e^{i2\pi W(\rho,\theta)/\lambda}\}\right|^{2}.
\end{equation}
Here $\mathcal{F}$ denotes the Fraunhofer diffraction operator \cite{Goodman2005}, $P(\rho,\theta)$ is the complex pupil function, $W(\rho,\theta)$ is the wavefront, and $\lambda$ is the wavelength \cite{Born1999}. Because of inherent imperfections such as diffraction limits \cite{Goodman2005}, manufacturing tolerances \cite{Smith2007}, assembly misalignment \cite{Mahajan1998}, and unavoidable camera shake and object motion \cite{Whyte2010,Hirsch2011}, the captured images are often corrupted by optical aberrations \cite{Mahajan1998,Noll1976} and motion blur \cite{Whyte2010,Sun2015}.
 
A straightforward way to improve image quality is to explicitly calibrate the system’s PSF by measuring optical aberrations and motion-blur kernels for each lens \cite{PlattShack2001,Paxman1992,Shaw1991}. However, this calibration process is often time-consuming, requires dedicated hardware, and relies heavily on expert knowledge.
%
Blind inverse problems offer a promising direction by jointly estimating both the latent image and the unknown degradation PSF \cite{Levin2009,Fergus2006}. This approach provides a unified framework for both calibration and deblurring, and has been applied to a wide range of applications not limited to deblurring, such as super-resolution and image denoising.

However, existing approaches often exhibit unstable convergence \cite{Levin2009}, limited prior expressiveness \cite{Krishnan2009}, and pronounced sensitivity to hyperparameters \cite{Cho2009,pan2019phase, al2025blind,laroche2024fast,ren2020neural}, which make them difficult to use reliably in practice.
%
Motivated by recent developments in self-diffusion \cite{luo2025selfdiffusionsolvinginverseproblems}, we propose DeblurSDI, a training-free and hyperparameter-insensitive blind imaging framework that achieves high stability while requiring no external priors.
%
Inheriting the core idea of self-diffusion, which progressively restores images across hierarchical noise levels, DeblurSDI extends this principle to jointly recover the sharp image and PSF. This formulation significantly improves robustness in the otherwise unstable joint optimization, particularly in PSF estimation, which is the primary challenge in blind imaging problems.

Extensive experiments on simulated optical aberrations and motion blur demonstrate that DeblurSDI consistently outperforms state-of-the-art blind deconvolution methods by a substantial margin. Our conditions are summarized as follows: 1) We propose DeblurSDI, a training-free and self-supervised framework for blind image recovery that integrates self-diffusion and jointly estimates both the sharp image and the PSF; 2) We construct evaluation datasets that include optical-aberration PSFs and motion-blur kernels, enabling comprehensive assessment of blind imaging performance under challenging conditions; 3) We demonstrate that DeblurSDI achieves consistently superior performance and robustness compared to state-of-the-art blind deblurring methods, particularly in stability and PSF estimation.

%% file: sec/2_Related_Work.tex
\section{Related Work}
\label{sec:related_work}

Correcting optical aberrations and motion blur generally follows one of two paradigms. In the model-based paradigm, the PSF is described using a parametric wavefront model such as Zernike polynomials \cite{Noll1976, Mahajan1994}, and calibration procedures are employed to estimate the corresponding coefficients. This yields an explicit PSF that enables non-blind deconvolution. In the model-free paradigm, blind deconvolution is performed, which bypasses calibration entirely by estimating both the image and the PSF from the observation \cite{ISO24157}.

\paragraph{Optical Aberration Modeling and PSF Calibration:}
Optical aberrations arise from diffraction limits \cite{Goodman2005}, manufacturing tolerances \cite{Smith2007}, lens-group misalignment \cite{Mahajan1998}, and environmental variations \cite{Born1999}, and they fundamentally shape the PSF of an imaging system. Early descriptions of aberrations were grounded in classical geometric optics \cite{Welford1986}, characterizing low-order distortions such as spherical aberration, coma, astigmatism, field curvature, and distortion \cite{Mahajan1998}. With the development of wave optics \cite{Goodman2005}, the PSF became formally modeled as the squared magnitude of the Fourier transform of the pupil function \cite{Hopkins1951}, where phase distortions encode the underlying wavefront errors. Modern optical systems typically represent these wavefront errors using Zernike polynomials \cite{ISO24157,Noll1976,Mahajan1994}, an orthonormal basis on the unit disk that directly corresponds to physically interpretable aberration modes and has become the industry standard for lens characterization \cite{ISO24157}.

Estimating the actual PSF of a physical lens, however, requires calibration procedures \cite{PlattShack2001,Paxman1992}. In controlled laboratory settings, wavefront sensors such as Shack–Hartmann devices \cite{PlattShack2001} provide direct measurements of Zernike coefficients, while phase-diversity methods infer aberrations from pairs of focused and defocused images \cite{Paxman1992,Gonsalves1982}. Microscopy systems often rely on submicron fluorescent beads \cite{Shaw1991} to sample spatially varying PSFs and construct eigen-PSF \cite{Schaub2019} or product-convolution models \cite{Aubert2022} for wide-field correction. Although accurate, these calibration pipelines require specialized hardware, multiple acquisitions, or carefully prepared samples, making them impractical for general-purpose cameras and consumer-level imaging devices.

As a result, industrial imaging systems such as smartphone cameras \cite{Chen2017HDR,Hasinoff2016}, surveillance sensors, and automotive optical modules—typically rely on one-time factory calibration combined with lookup tables \cite{Chen2017HDR,Wronski2019} or compact neural modules embedded in the ISP pipeline. These approaches correct only approximate or averaged aberrations and cannot account for camera shake, object motion, or per-instance PSF variations in real scenes. Therefore, while model-based PSF calibration offers high fidelity, its cost, hardware dependency, and lack of adaptability highlight the need for calibration-free alternatives such as blind deconvolution methods \cite{Levin2009,Fergus2006}.

\paragraph{Blind Deconvolution with Training-free or Trained Priors:}

Blind deconvolution aims to recover both the latent sharp image and the unknown blur kernel from a single degraded observation. Classical approaches typically formulate this as an alternating optimization problem that minimizes an energy functional over the image and the PSF \cite{Levin2009,Fergus2006}. Early methods rely on handcrafted priors such as sparsity, heavy-tailed gradient distributions, or total variation to regularize the ill-posed joint estimation \cite{Krishnan2009,Cho2009}. Although effective in limited settings, these methods often exhibit strong sensitivity to initialization, kernel size, and hyperparameter choices, which leads to kernel drift and unstable convergence behaviors.

To alleviate the limitations of handcrafted priors, implicit neural priors have emerged as a powerful alternative. Deep Image Prior (DIP) demonstrates that the structure of an untrained convolutional network can serve as a natural image prior, enabling image restoration without external data \cite{Ulyanov2018DIP}. Subsequent works extend this idea to blind deblurring by optimizing both the image and the kernel through coupled implicit networks \cite{ren2020neural}. These methods offer improved flexibility and avoid the need for large training datasets, yet the coupled optimization is still highly unstable and prone to overfitting, especially when the blur kernel is large or spatially complex.

Another line of work introduces pretrained generative models such as GANs, VAEs, and more recently diffusion models as strong image priors for inverse problems \cite{Bigdeli2017,Gandelsman2019,Whang2022DDRM}. Diffusion-based solvers, including diffusion posterior sampling (DPS) and related score-based formulations, have demonstrated impressive performance in non-blind deblurring and other linear inverse problems \cite{Chung2023DPS,Kawar2023DPS}. However, these methods rely heavily on large-scale pre-training and are restricted by domain mismatch between the training data and the target scene. More importantly, pretrained-prior approaches are generally incapable of estimating the blur kernel, and therefore cannot be directly applied to fully blind deconvolution.

Because existing methods either require dedicated calibration hardware or struggle with the inherent instability of joint image–kernel estimation. 
These limitations highlight the need for a framework that can both capture realistic optical degradations and robustly estimate the PSF without external training or calibration. Motivated by this gap, we now introduce our degradation modeling pipeline and the proposed DeblurSDI framework.

%% file: sec/3_method.tex
\section{Method}
\label{sec:method}

In this section, we first introduce our aberration simulation framework, which generates a family of physically realistic PSFs derived from wavefront distortions modeled by Zernike polynomials. These simulated PSFs allow us to systematically evaluate blind imaging performance under controlled optical conditions. We then present DeblurSDI, a training-free blind deblurring framework that jointly recover the latent sharp image and the unknown PSF. By combining physically grounded PSF synthesis with a robust blind inverse solver, our approach enables comprehensive analysis and effective restoration across a wide range of optical aberrations and motion blur scenarios.

\subsection{Modeling Optical Aberrations}
\label{subsec:optical_aberrations}
To simulate realistic optical degradations, we adopt the standard wavefront–pupil–PSF formulation widely used in computational optics. Optical aberrations are represented through a weighted sum of Zernike polynomials \cite{Mahajan1994}, which form an orthonormal basis on the unit disk and are known to accurately model low- and high-order aberration modes such as defocus, astigmatism, coma, spherical aberration, trefoil, and quadrafoil.

\paragraph{Zernike-based wavefront modeling:}
Let $(\rho,\theta)$ denote polar coordinates on the normalized pupil domain
$\{(x,y): x^2+y^2\le 1\}$.
The Zernike polynomial of radial order $n$ and azimuthal order $m$ is 
defined as
\begin{equation}
Z_n^{m}(\rho,\theta) =
\begin{cases}
R_n^{|m|}(\rho)\cos(m\theta), & m \ge 0,\\[3pt]
R_n^{|m|}(\rho)\sin(|m|\theta), & m < 0,\\
\end{cases}
\end{equation}
where the radial term $R_{n}^{m}(\rho)$ is given by
\begin{equation}
R_n^{m}(\rho) =
\sum_{k=0}^{\frac{n-m}{2}}
(-1)^k\,
\frac{(n-k)!}
{k!\,
\big(\tfrac{n+m}{2}-k\big)!\,
\big(\tfrac{n-m}{2}-k\big)!}\,
\rho^{\,n-2k}.
\end{equation}

A wavefront corrupted by multiple aberrations can therefore be written as
\begin{equation}
W(\rho,\theta)
= \sum_{(n,m)\in\mathcal{A}} 
a_{n,m} \, Z_n^{m}(\rho,\theta),
\label{eq:wavefront}
\end{equation}
where each coefficient $a_{n,m}$ controls the severity of the corresponding
aberration mode. In our simulations, we include all Zernike modes up to
order $n=4$ (defocus, astigmatism, coma, trefoil, spherical, quadrafoil),
which are sufficient to reproduce a wide range of realistic aberration
patterns observed in lens-based imaging systems.

\paragraph{Pupil function and PSF generation:}
Given the aberrated wavefront $W$, the complex pupil function is
\begin{equation}
P(\rho,\theta)
= \mathbf{1}_{\{\rho\le 1\}}\,
\exp\!\left(\frac{2\pi i}{\lambda}\,W(\rho,\theta)\right),
\label{eq:pupil}
\end{equation}
where $\lambda$ denotes the wavelength and the indicator ensures that
the phase modulation is applied only within the physical aperture.
The PSF is obtained as the squared
magnitude of the Fourier transform of the pupil:
\begin{equation}
h(x,y)
= \frac{\left|\mathcal{F}\!\left\{P(\rho,\theta)\right\}(x,y)\right|^2}
     {\max\limits_{x,y}
      \left|\mathcal{F}\!\left\{P(\rho,\theta)\right\}(x,y)\right|^2},
\label{eq:psf}
\end{equation}
where normalization ensures that $\max h = 1$.
We use an odd-sized Fourier grid so that the PSF peak is centered exactly at the middle pixel. A central crop of $h$ is used as the final convolution PSF in all experiments.

\subsection{Blind Imaging via Self-Diffusion}
This section introduces the DeblurSDI method, which addresses blind optical correction by extending the {self-diffusion} framework \cite{luo2025selfdiffusionsolvinginverseproblems}. The original self-diffusion method was designed for non-blind problems where the degradation operator is known. Our key contribution is to adapt this powerful framework to the more challenging blind imaging setting, where the degradation operator is also unknown. To achieve this, we introduce a novel process that jointly recovers the clean image and the PSF by optimizing two coupled, untrained neural networks within a self-contained reverse diffusion process.

\paragraph{Self-Diffusion:}

Self-diffusion is a pretraining free method designed to solve general linear inverse problems of the form $\mathcal{A}\mathbf{x}_{\text{true}} = \mathbf{y}$, where $\mathcal{A}$ is a known forward operator, $\mathbf{x}_{\text{true}}$ is the unknown solution, and $\mathbf{y}$ is the observation. It operates via an iterative reverse diffusion process that starts from pure Gaussian noise. At each step $t$, a noisy version of the current estimate $\mathbf{x}_t$ is created with
\begin{equation}
    \hat{\mathbf{x}}_t = \mathbf{x}_t + \sigma_t \cdot \epsilon_t ~.
\end{equation}
A single, randomly initialized network at the first time step, the self-denoiser $D_\theta$, is then optimized continuously by minimizing a data fidelity loss with respect to the original observation $\mathbf{y}$ for each time step $t$,
\begin{equation}
    \mathcal{L}_t(\theta) = \| \mathcal{A} D_{\theta,t}(\hat{\mathbf{x}}_t) - \mathbf{y} \|_2^2~.
\end{equation}
The effectiveness of this process stems from a principle known as noise-regulated spectral bias. The noise schedule $\sigma_t$ implicitly regularizes the optimization, forcing the network to first learn low-frequency components and progressively refine high-frequency details in a multi-scale manner.

\paragraph{Joint Image and PSF Estimation:}
Based on above observations, we apply such implicit regularization to PSF estimation. The forward model for blind image deblurring is given by
\begin{equation}
    \mathbf{y} = \mathbf{x}_\text{true} \circledast \mathbf{k} + n~,
\end{equation}
where the $\circledast$ denotes convolution, the sharp image $\mathbf{x}_\text{true} \in \mathbb{R}^{H \times W \times C}$ and the PSF $\mathbf{k} \in \mathbb{R}^{K \times K \times 1}$ are both unknown. To adapt the self-diffusion framework to this blind setting, both variables must be estimated simultaneously. We achieve this by employing two dedicated, randomly initialized networks: an image denoiser $D_\theta$ to restore $\mathbf{x}_\text{true}$, and a PSF generator $G_\phi$ to produce $\mathbf{k}$.
Our method simulates a reverse diffusion process over $T$ discrete time steps, starting with random noise for both the image estimate, $\mathbf{x}_T$, and the PSF estimate, $\mathbf{z}_T$. At each time step $t \in \{T, T-1, ..., 1\}$, the current estimates are perturbed with scheduled noise,
\begin{equation}
    \hat{\mathbf{x}}_t = \mathbf{x}_t + \sigma_t \cdot \epsilon_x, \quad \text{and} \quad \hat{\mathbf{z}}_t = \mathbf{z}_t + \sigma'_t \cdot \epsilon_z,
\end{equation}
where $\epsilon_x \sim \mathcal{N}(0, \mathbf{I})$ and $\epsilon_z \sim \mathcal{N}(0, \mathbf{I})$. The noise schedule is \( \sigma_t = \sqrt{1 - \bar{\alpha}_t},~\text{where}~ \bar{\alpha}_t = \mathop{\textstyle\prod}\nolimits_{i=0}^t (1 - \beta_i)~\text{and}~\beta_t = \beta_{\text{end}} + \frac{t}{T - 1}(\beta_{\text{start}} - \beta_{\text{end}})\), and $\sigma'_t = \mu\sigma_t, \mu$ is a adjutable hyperparameter.
It is often known that the PSF is to be sparse in practice. We thus define $R(\cdot)$ as L1-norm and use many layers of ReLU in the PSF generator to enforce sparsity on the generated PSF. The both networks are jointly optimized within an inner loop by minimizing the following objective:
\begin{equation}
    \mathcal{L}_t(\theta, \phi) = {\| (D_\theta(\hat{\mathbf{x}}_t) \circledast G_\phi(\hat{\mathbf{z}}_t)) - \mathbf{y} \|_2^2} + {\lambda_k {R}(G_\phi(\hat{\mathbf{z}}_t))}   
\end{equation}
After the inner optimization loop, the improved networks produce cleaner estimates for the next time step, $\mathbf{x}_{t-1} = D_\theta(\hat{\mathbf{x}}_t)$ and $\mathbf{z}_{t-1} = G_\phi(\hat{\mathbf{z}}_t)$, continuing the coarse-to-fine reconstruction inherent to the self-diffusion process. The detailed algorithm is presented in Algorithm~\ref{alg:gk-sdi}.
\begin{algorithm}[t]
    \caption{Blind Imaging via Self-Diffusion}
    \label{alg:gk-sdi}
    \begin{algorithmic}[1]
        \Require Blurry image $\mathbf{y}$, total steps $T$, inner iterations $S$, learning rate $\eta$, $\ell_1$ weights $\lambda_k$
        \State \textbf{Initialize:}
        \State Image estimate $\mathbf{x}_T \sim \mathcal{N}(0, \mathbf{I})$; $D_\theta$ with random weights $\theta$
        \State PSF estimate $\mathbf{z}_T \sim \mathcal{N}(0, \mathbf{I})$; $G_\phi$ with random weights $\phi$
        \State Adam optimizer for $(\theta, \phi)$
        \State Noise schedule $\sigma_t$ for $t \in \{1, ..., T\}$
        \For{$t = T, T-1, ..., 1$}
            \State Sample noise $\epsilon_\mathbf{x}, \epsilon_\mathbf{z} \sim \mathcal{N}(0, \mathbf{I})$
            \State Create noisy inputs: $\hat{\mathbf{x}}_t \leftarrow \mathbf{x}_t + \sigma_t \cdot \epsilon_\mathbf{x}$, and $\hat{\mathbf{z}}_t \leftarrow \mathbf{z}_t + \sigma'_t \cdot \epsilon_\mathbf{z}$
            \For{$s = 1, ..., S$}
                \State Generate PSF: $\mathbf{k} \leftarrow G_\phi(\hat{\mathbf{z}}_t)$
                \State Compute denoised image: $\mathbf{x}_{t} \leftarrow D_\theta(\hat{\mathbf{x}}_t)$
                \State Calculate loss: $\mathcal{L}(\theta, \phi) \leftarrow \| (\mathbf{x}_{t} \circledast \mathbf{k}) - \mathbf{y} \|_2^2 + \lambda_k {R}(\mathbf{k})$
                \State Update parameters via gradient descent: $(\theta, \phi) \leftarrow (\theta, \phi) - \eta \nabla_{(\theta, \phi)} \mathcal{L}(\theta, \phi)$
            \EndFor
            \State Update image estimate: $\mathbf{x}_{t-1} \leftarrow D_\theta(\hat{\mathbf{x}}_t)$
            \State Update PSF latent code: $\mathbf{z}_{t-1} \leftarrow G_\phi(\hat{\mathbf{z}}_t)$
        \EndFor
        \State \textbf{return} Reconstructed image $\mathbf{x}\leftarrow\mathbf{x}_0$ and final PSF $\mathbf{k}\leftarrow\mathbf{k}_0$
    \end{algorithmic}
\end{algorithm}

%% file: sec/4_Exps.tex
\section{Implementation and Experiments}
\label{sec:exps}

\subsection{Dataset}
\label{sec:dataset}
\begin{figure}[h]
    \centering
    \includegraphics[width=1.0\linewidth]{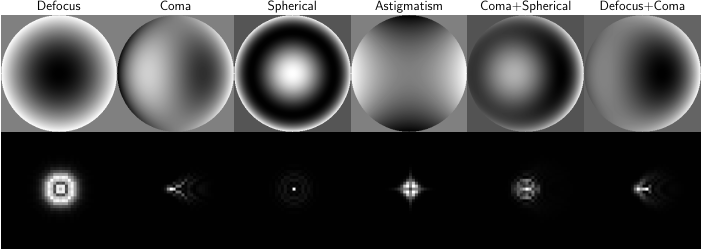}
    \vspace{-15pt}
    \caption{PSFs for individual and combined aberrations (defocus, coma, astigmatism, spherical, coma+spherical, defocus+coma). Uprow: wavefronts. Downrow: PSFs.}
    \label{fig:aberrations}
\end{figure}

\paragraph{Optical Aberration:}
The simulated optical aberration process is implemented numerically by evaluating the Zernike
basis on a $255\times255$ Cartesian grid, constructing the pupil
(\ref{eq:pupil}), and computing the PSF via FFT according to
(\ref{eq:psf}).  
Figure~\ref{fig:aberrations} shows the resulting wavefronts and PSFs for individual and combined aberrations (defocus, coma, astigmatism, spherical, coma+spherical, defocus+coma). These aberrations are all applied on sharp images collected from above datasets to obtain blurred images. Then our proposed DeblurSDI is applied and evaluated on them.

\paragraph{Motion Blur:}
To systematically evaluate the performance of our method, we collected four datasets including Levin \citep{levin2007image}, Cho \citep{cho2009fast}, Kohler \citep{kohler2012recording}, and FFHQ \citep{karras2019style}. The first three datasets are widely used benchmarks for blind image deblurring, containing various synthetic blur kernels applied to natural images. The FFHQ dataset consists of 20 random selected human face images and 4 blur kernels from Cho \citep{cho2009fast}. For our experiments, we use $T=30$ outer reverse diffusion steps. At each step, we run $S=200$ inner optimization iterations.
For each dataset, we use the provided blur kernels and generate blurred images by convolving the sharp images with these kernels to simulate real-world conditions. Each blur kernel is applied to every image in the dataset. The details of our comprehensive datasets are shown in Table~\ref{table:datasets}.
\begin{table}[h]
\centering
\caption{Details of datasets for evaluation}
\vspace{-10pt}
\resizebox{\columnwidth}{!}{
\begin{tabular}{cccc}
\toprule
Dataset & Image Size & Kernel Sizes \\ 
\midrule

Levin\textsuperscript{1} &
255$\times$255 &
\{19, 17, 15, 27, 13, 21, 23, 23\}
\\

Cho\textsuperscript{2} &
\begin{tabular}{@{}c@{}}
\footnotesize 622$\times$463, 780$\times$580,\\
\footnotesize 1006$\times$665, 1002$\times$661
\end{tabular} &
\{27, 23, 19, 21\}
\\

Kohler\textsuperscript{3} &
800$\times$800 &
\begin{tabular}{@{}c@{}}
16, 14, 9, 13, 29, 17, 19, 98,\\
102, 62, 40, 29
\end{tabular}
\\

FFHQ\textsuperscript{4} &
256$\times$256 &
\{27, 23, 19, 21\}
\\

\midrule
\multicolumn{2}{c}{Total Pairs} & 128 \\
\bottomrule
\end{tabular}}
\footnotesize
\label{table:datasets}
\end{table}

\subsection{Network architecture and training}
Due to the low-dimensional nature of the blur kernel, we employ a fully-connected network (FCN) to implement the kernel generator, $G_\phi$. To ensure the output corresponds to a physical blur kernel, a softmax activation is applied to the final layer, enforcing non-negativity and a sum-to-one constraint. The 1D output of $G_\phi$ is subsequently reshaped into a 2D blur kernel. Besides, we introduced standard mode for ablation study where the latent vector $z$ is sampled from a normal distribution and kept fixed during training, and diffusion mode where the $\mathbf{z}_t$ evolves through the self-diffusion process. $K$ is the kernel size, $n$ is the number of hidden layers and $H_d$ is the hidden dimension size, Table \ref{tab:arch_g_phi} shows the architecture in details. 

\begin{table}[h]
\centering
\caption{The architecture of the kernel generator $G_\phi$}
\vspace{-10pt}
\label{tab:arch_g_phi}
\resizebox{\columnwidth}{!}{
\begin{tabular}{lll}
\toprule
\textbf{Mode} & \textbf{Layer} & \textbf{Specification } \\ \midrule
\multirow{3}{*}{\begin{tabular}[c]{@{}l@{}}{Standard}\\ \end{tabular}} & Input & $z \in \mathbb{R}^{200}\sim \mathcal{N}(0, \mathbf{I})$\\ 
 & Hidden layer & Linear(200, 2000); ReLU6 \\ 
 & Output layer & Linear(2000, $K \times K$); Softmax \\ \midrule
\multirow{5}{*}{\begin{tabular}[c]{@{}l@{}}{Diffusion}\\ \end{tabular}} & Input & $\mathbf{z}_t \in \mathbb{R}^{K \times K}$ \\ 
 & Hidden layer 1 & Linear($K \times K$, $H_d$); ReLU \\ 
 & Hidden layer 2 & Linear($H_d$, $H_d$); ReLU \\ 
 & \dots & \dots \\ 
 & Hidden layer $n$ & Linear($H_d$, $H_d$); ReLU \\ 
 & Output layer & Linear($H_d$, $K \times K$); Softmax \\ \bottomrule
\end{tabular}
}
\end{table}
For the image denoiser $D_\theta$, we employ an encoder-decoder network with skip connections, following a U-Net-like structure. The network consists of five hierarchical levels. Each level in the encoder path consists of two convolutional blocks and a stride-2 convolution for downsampling. Correspondingly, the decoder path uses bilinear upsampling. Skip connections concatenate features from each encoder level to the corresponding decoder level. Non-Local Blocks are integrated into the deeper encoder levels (levels 3, 4, and 5) to capture long-range dependencies.
The architecture is detailed as in Table \ref{tab:arch_d_theta}.

\begin{table}[H]
\centering
\caption{The architecture of the image denoiser $D_\theta$, consisting of five encoder units ($e_i$) and five decoder units ($d_i$). The form is Conv(input channels, output channels, kernel size).}
\vspace{-10pt}
\label{tab:arch_d_theta}
\begin{tabular}{ll}
\toprule
\textbf{Layer} & \textbf{Specification} \\ \midrule
\textbf{Input} & Noisy image $\hat{x}_t \in \mathbb{R}^{C \times H \times W}$ \\ 
\textbf{Output} & Denoised image $x_{t-1} \in \mathbb{R}^{C \times H \times W}$ \\ 
Encoder unit 1 & $e_1(\cdot, 128, 3)$ \\ 
Encoder unit 2 & $e_2(128, 128, 3)$ \\ 
\multicolumn{2}{c}{$\cdots$}  \\
Encoder unit 5 & $e_5(128, 128, 3)$ \\ 
Decoder unit 5 & $d_5(128, 128, 3)$ \\ 
\multicolumn{2}{c}{$\cdots$}  \\
Decoder unit 2 & $d_2(128, 128, 3)$ \\ 
Decoder unit 1 & $d_1(128, 128, 3)$ \\ 
Output layer & Conv(128, C, 1); Sigmoid \\ \bottomrule
\end{tabular}
\end{table}

We use a single Adam optimizer to jointly update the parameters of both the image denoiser, $D_\theta$, and the kernel generator, $G_\phi$.
The initial learning rate for the image denoiser $D_\theta$ is set to $1 \times 10^{-3}$. The kernel generator $G_\phi$ uses a lower learning rate, typically 25\% of the denoiser's rate (i.e., $2.5 \times 10^{-4}$), as the small change in the kernel can make a bigger impact on the image. The smaller learning rate helps stable convergence of the kernel estimate.
Furthermore, we employ an optional adaptive learning rate schedule for the kernel generator. The learning rate is decayed by a factor of 0.95 at the end of each outer time step $t$, down to a minimum threshold of $1 \times 10^{-5}$.
The L1 regularization weight for the kernel prior is set to $\lambda_k = 2 \times 10^{-3}$.
The noise level $\sigma_t$ for the image perturbation at each step $t$ is determined by a pre-defined variance schedule. Following common practice in diffusion models, we use a linear schedule where the variance $\beta_t$ interpolates from $\beta_{\text{start}} = 1 \times 10^{-4}$ to $\beta_{\text{end}} = 2 \times 10^{-2}$ over $T$ steps. The noise level $\sigma_t$ is then derived from the cumulative product of these variances.

\subsection{Experiments}
In the following, we present the deblurring process described in \Cref{alg:gk-sdi} and evaluated DeblurSDI from different perspectives, including robustness to the choice of PSF generator, sensitivity to hyperparameters, and noise degradation.
\begin{figure*}[t]
\centering
\includegraphics[width=0.9\textwidth]{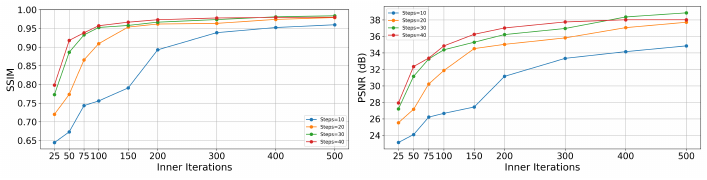}
\vspace{-13pt}
\caption{Sensitivity of noise scheduled steps, $T$, and inner iterations, $S$. The graphs show the SSIM (left) and PSNR (right) scores for different numbers of outer diffusion steps ($T \in \{10, 20, 30, 40\}$) and inner optimization iterations ($S \in \{25, ..., 500\}$).}
\label{fig:steps_n_iters}
\end{figure*}
\vspace{-0.5em}
\paragraph{Hyperparameters sensitivity:}
The number of outer diffusion steps $T$ and inner optimization iterations $S$ in \Cref{alg:gk-sdi} are two critical hyperparameters that directly impact both the reconstruction quality and computational cost. Intuitively, more diffusion steps allow for a finer coarse-to-fine reconstruction, while more inner iterations enable better convergence of the networks at each step. However, increasing either parameter also leads to longer runtimes.
To evaluate the sensitivity of our method to these parameters, we conduct experiments varying $T$ from 10 to 40 and $S$ from 25 to 500. As shown in Figure~\ref{fig:steps_n_iters}, performance improves with higher values of $T$ and $S$. The most significant gains occur when increasing $T$ from 10 to 30. However, the improvements tend to saturate beyond certain thresholds (e.g., $T=30$, $S=400$), with the performance curve for $T=40$ closely tracking that of $T=30$. This indicates that our approach can achieve strong deblurring performance without requiring excessively high iteration counts.
\vspace{-0.5em}

\begin{figure}[h]
    \centering
    \includegraphics[width=\columnwidth]{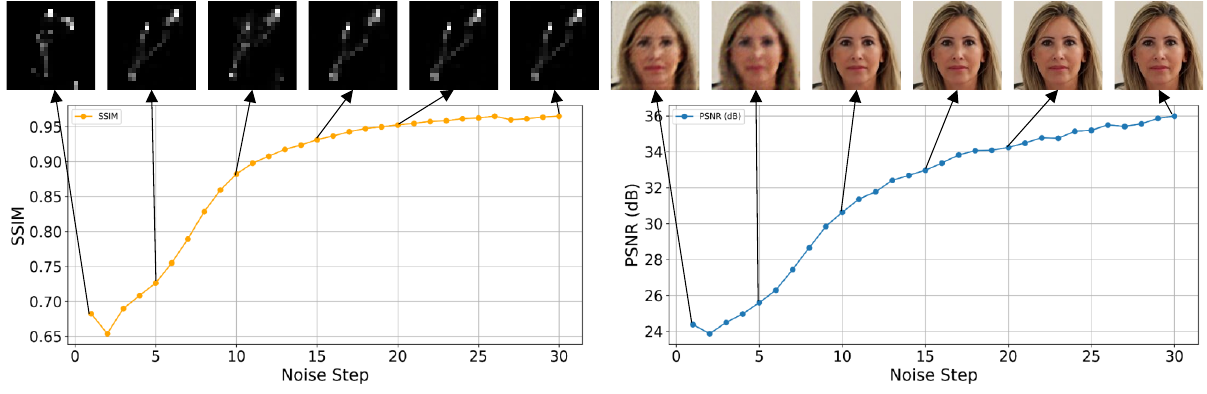}
    \vspace{-20pt}
    \caption{{Evolution of image and kernel estimates during DeblurSDI's reverse diffusion process.}}
    \label{fig:deblurring_process}
\end{figure}

\paragraph{Recovering process:}
Figure \ref{fig:deblurring_process} shows the evolution of estimates of image and kernel through the deblurring process. The left and right subfigures shows the SSIM and PSNR between the original image and the reconstruction over noise steps. The estimates of images and kernels at noise steps $5, 10, 15, 20, 30$ are shown on the top. Unlike traditional optimization processes where evaluation metrics typically increase monotonically, our curves exhibit an up-down-up behavior (especially for PSNR curve), which is attributed to the noise scheduling strategy. By injecting noise into intermediate reconstruction results, we effectively enlarge the search space of the inverse solution. The initial reconstructions are smooth and lack fine detail, while later steps recover sharper features. This aligns with the coarse-to-fine nature of self-diffusion.\vspace{-0.75em}
\vspace{-2em}

\begin{figure}[H]
    \centering
    \includegraphics[width=\columnwidth]{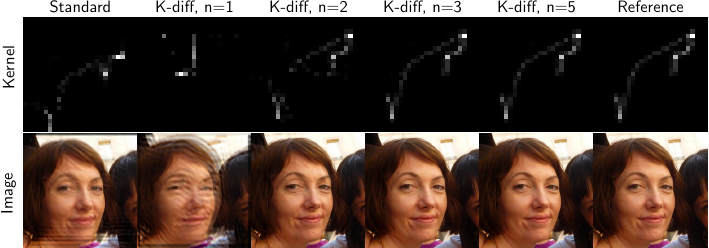}
    \vspace{-15pt}
    \caption{Ablaion study on the network for PSF estimation. This figure compares the performance of the ``Standard" mode against the ``Diffusion" mode architecture with a varying number of hidden layers ($n=1$ to $n=5$).}
    \label{fig:diff-std}
\end{figure}

\paragraph{Architecture of kernel generator:}
Figure \ref{fig:diff-std} shows how the architecture of PSF kernel generator $G_\phi$ affects the performance of DeblurSDI. The ``Standard" mode is $G_\phi$ without noise schedule, while ``k-diff'' stands for the kernel generator $G_\phi$ with noise schedule as Table \ref{tab:arch_g_phi} describes. The results show that the ``k-diff" mode outperforms the ``Standard" mode in terms of PSNR and SSIM. The results indicate that increasing the depth of the kernel generator network leads to a more accurate kernel estimation.\vspace{-0.25em}

\begin{figure}[H]
    \centering
    \includegraphics[width=\linewidth]{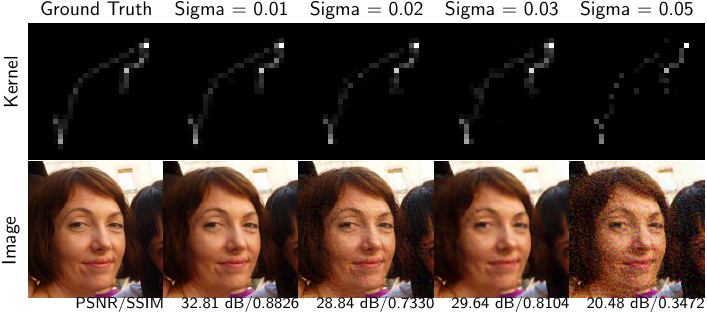}
    \vspace{-20pt}
    \caption{Performance comparison under different addictive noise on the blurred image.}
    \label{fig:results_noise}
\end{figure}

\vspace{-3em}
\paragraph{Noise degradation:}
To evaluate the robustness of our method to additive noise, we test it across noise levels $\sigma \in {0.01, 0.02, 0.03, 0.05}$ using the degradation model $\mathbf{y} = \mathbf{x} \circledast \mathbf{k} + \sigma \mathbf{n}, ~ \mathbf{n} \sim \mathcal{N}(0, \mathbf{I}).
$ The image is normalized to $[0,1]$. As shown in Figure~\ref{fig:results_noise}, our method remains stable for noise levels up to $\sigma = 0.03$, achieving PSNR above $28.0$ and SSIM above $0.73$. The recovered PSFs also remain visually accurate up to $\sigma = 0.03$.\vspace{-0.5em}

%% file: sec/5_Evaluation.tex
\section{Evaluation}
\label{sec:evaluation}
\begin{figure*}[t]
    \centering
    \includegraphics[width=0.9\linewidth]{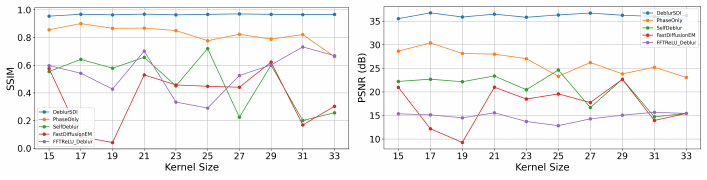}
    \vspace{-12pt}
    \caption{Performance and stability comparison across different kernel sizes. The graphs show the SSIM (left) and PSNR (right) scores for five deblurring methods evaluated on kernel sizes ranging from 15 to 33. Our method, DeblurSDI (blue), consistently achieves the highest scores and demonstrates remarkable stability, with its performance remaining largely unaffected by changes in kernel size. In contrast, other methods exhibit significant volatility, underscoring the superior robustness of our approach.}
    \label{fig:results_kernelsize}
\end{figure*}
In this section, we evaluated DeblurSDI on datasets described in \Cref{sec:dataset} comparing to other methods. The joint estimation of the image denoiser and the blur kernel often collapses to trivial solutions, such as Dirac kernels or reproducing the blurred image itself, especially when the chosen kernel size is incompatible with the image content. Larger kernels are harder to be recovered accurately, while smaller kernels may fail to capture long-range motion. For this reason, SelfDeblur \citep{ren2020neural} carefully selects the kernel size for each image.
In contrast, our method exhibits much greater robustness. As shown in Figure~\ref{fig:results_kernelsize}, we evaluate ten different kernel sizes from 15 to 33, and compare the performance of several approaches. Our method not only achieves consistently superior performance across all kernel sizes but also demonstrates remarkable stability.

\begin{figure}[H]
\centering
\includegraphics[width=1\linewidth]{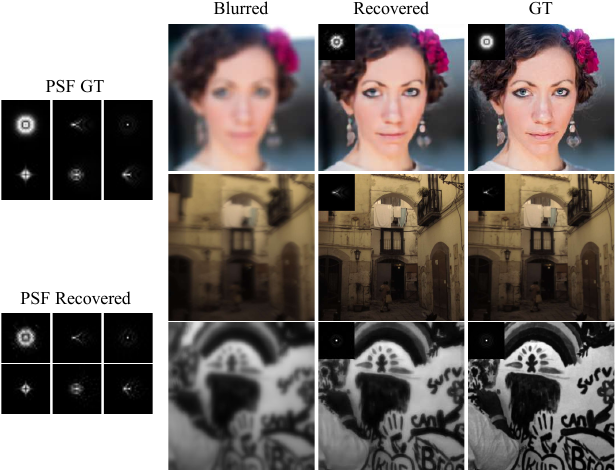}
\vspace{-15pt}
\caption{The reconstruction results of DeblurSDI on the optical aberration situations.}
\label{fig:results_aberration}
\end{figure}

\paragraph{Optical Aberration Correction:}
Based on the simulated six optical aberrations from \ref{subsec:optical_aberrations}, all the clean images are blurred by each one of them to obtain the corresponding blurred images. Then the proposed DeblurSDI and all the compared methods are applied and evaluated on the blurred images. The results are shown in Figure \ref{fig:results_aberration} and the quantitative results are shown in Table \ref{tab:quantitative_results_aberration}. It can be seen that the proposed DeblurSDI outperforms all the compared methods in all the optical aberration situations.

\begin{table}[h]
\centering
\caption{\footnotesize Quantitative results of optical aberration correction performance (PSNR/SSIM) on four datasets.}
\vspace{-10pt}
\label{tab:quantitative_results_aberration}
\resizebox{\columnwidth}{!}{
\begin{tabular}{lccccc}
\toprule
 & Phase-Only\citep{pan2019phase} & FFT-ReLU Deblur\citep{al2025blind} & SelfDeblur\citep{ren2020neural} & FastDiffusionEM\citep{laroche2024fast} & DeblurSDI (Ours) \\ \midrule
Levin\citep{levin2007image}    & 15.52/0.3722 & 19.57/0.5655 & 18.13/0.4706 & 18.68/0.5085 & 28.36/0.8598 \\
Kohler\citep{kohler2012recording}   & 27.37/0.7889 & 29.89/0.8358 & 20.76/0.5409 & 19.83/0.5242 & 32.07/0.9061 \\
FFHQ\citep{karras2019style}     & 26.31/0.7703 & 23.21/0.6942 & 19.65/0.5591 & 17.90/0.4508 & 33.00/0.9343 \\ \bottomrule
\end{tabular}
}
\end{table}
\vspace{-1em}

\begin{figure}[h]
\centering
\includegraphics[width=1\linewidth]{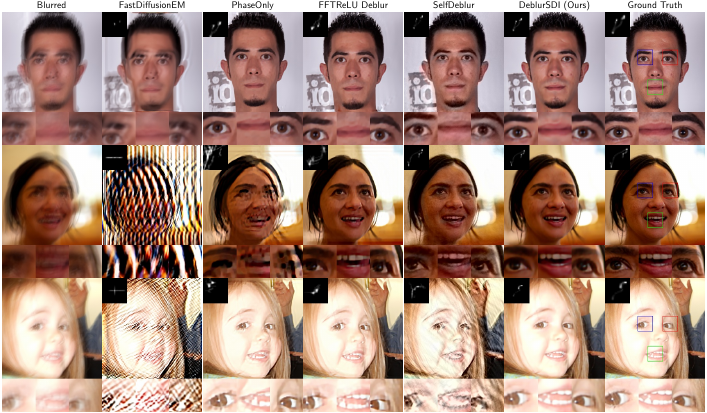}
\vspace{-15pt}
\caption{Deblurring results on the FFHQ dataset \citep{karras2019style}.}
\label{fig:results_ffhq}
\end{figure}

\paragraph{Motion Deblur:}
Figure \ref{fig:results_ffhq} shows deblurring results of different methods on the FFHQ dataset \citep{karras2019style}.
For each estimated image, the recovered kernel is displayed at the top-left corner, and three zoomed-in regions highlight fine details. 
As observed, FastDiffusionEM \citep{laroche2024fast} performs the worst. 
Despite being pre-trained on the FFHQ dataset, its performance remains poor: the estimated kernels degenerate into trivial point- or line-like structures. 
Without reliable kernel estimation, even a strong image prior cannot yield satisfactory deblurring. 
In contrast, other four optimization-based methods produce visibly superior results.

\begin{figure}[t]
\centering
\includegraphics[width=1\linewidth]{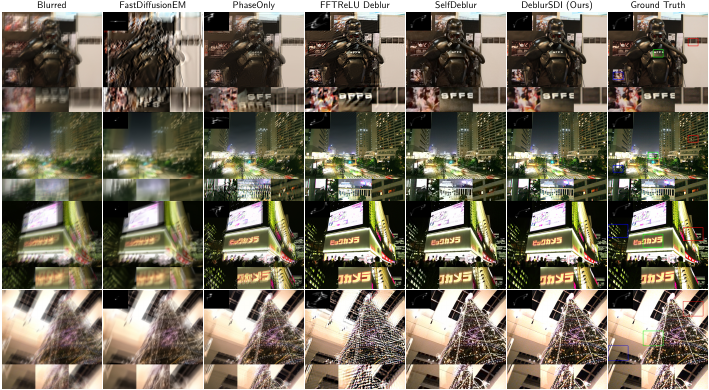}
\vspace{-15pt}
\caption{Deblurring results on the Cho dataset \citep{cho2009fast}.}
\label{fig:results_cho}
\end{figure}

\begin{figure}[t]
\centering
\includegraphics[width=1\linewidth]{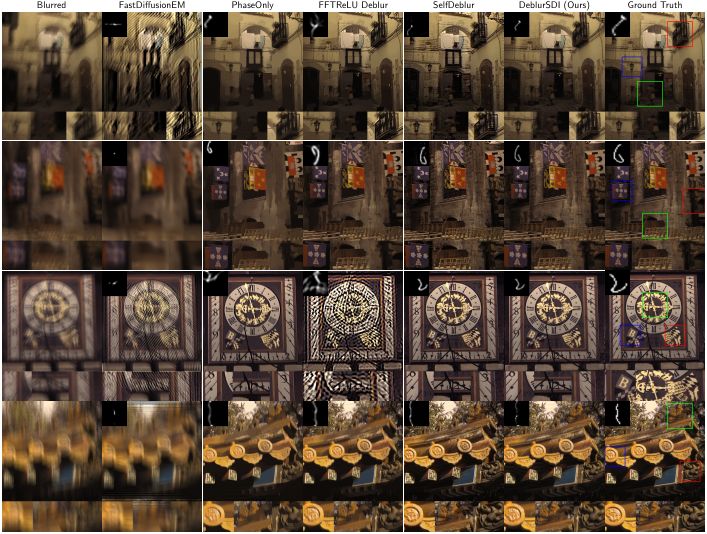}
\vspace{-15pt}
\caption{Deblurring results on the Kohler dataset \citep{kohler2012recording}.}
\label{fig:results_kohler}
\end{figure}

\begin{figure}[t]
\centering
\includegraphics[width=1\linewidth]{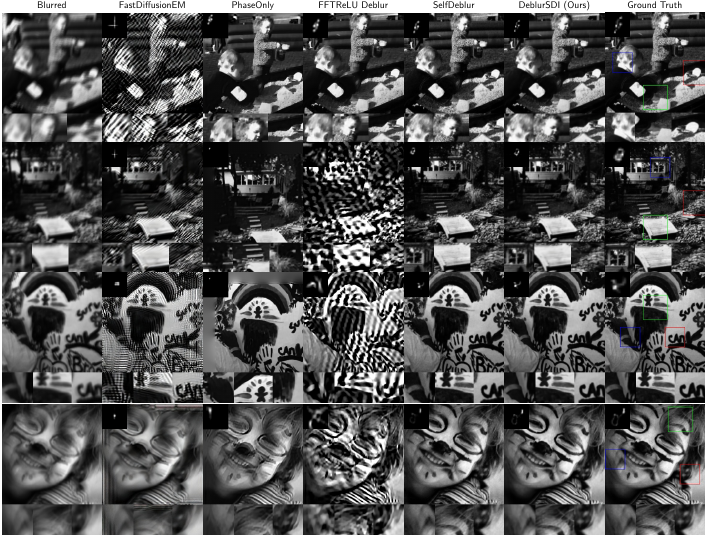}
\vspace{-15pt}
\caption{Deblurring results on the Levin dataset \citep{levin2007image}.}
\label{fig:results_levin}
\end{figure}

PhaseOnly \citep{pan2019phase} and FFTReLU Deblur \citep{al2025blind} provide good results in some cases, but lack of robustness when facing various blur kernels. SelfDeblur \citep{ren2020neural} shows the most promising performance out of previous methods, yet still suffers from poor generalizability, reconstruction shiftting and color distortion. The proposed DeblurSDI method consistently outperforms all compared methods, achieving significant improvements in the long-standing shiftting and robustness issues of blind deblurring techniques. Moreover, DeblurSDI is also able to recover accurate blur kernels, which gurantees the performance of the method. More results of other datasets are shown in  Figure~\ref{fig:results_cho}, \ref{fig:results_kohler}, and \ref{fig:results_levin}.

\begin{table}[h]
\centering
\caption{\footnotesize Quantitative results of blind motion deblurring performance (PSNR/SSIM).}
\vspace{-10pt}
\label{tab:quantitative_results}
\resizebox{\linewidth}{!}{   
\begin{tabular}{lccccc}
\toprule
 & Phase-Only\citep{pan2019phase} & FFT-ReLU Deblur\citep{al2025blind} &
 SelfDeblur\citep{ren2020neural} & FastDiffusionEM\citep{laroche2024fast} & DeblurSDI (Ours) \\ \midrule
Levin\citep{levin2007image}  & 20.68/0.6061 & 15.56/0.3845 & 25.06/0.7301 & 16.55/0.4005 & 31.85/0.7911 \\
Cho\citep{cho2009fast}    & 19.89/0.6746 & 18.73/0.6546 & 20.37/0.6844 & 15.39/0.4687 & 28.73/0.8859 \\
Kohler\citep{kohler2012recording} & 28.23/0.8092 & 25.33/0.7140 & 21.97/0.5995 & 18.85/0.4813 & 29.17/0.7653 \\
FFHQ\citep{karras2019style}   & 25.80/0.7904 & 21.71/0.6579 & 19.82/0.5563 & 15.59/0.3592 & 33.90/0.9064 \\
\bottomrule
\end{tabular}
}
\end{table}

We compared our DeblurSDI method with several other blind deblurring approaches, including Phase-Only \citep{pan2019phase}, FFT-ReLU Deblur \citep{al2025blind}, SelfDeblur \citep{ren2020neural}, and FastDiffusionEM \citep{laroche2024fast} and calculated Peak Signal-to-Noise Ratio (PSNR) and Structural Similarity Index Measure (SSIM) on each dataset. The quantitative results are summarized in Table~\ref{tab:quantitative_results}. This demonstrates the effectiveness and generalizability of our self-diffusion approach in recovering sharp images and accurate blur kernels.

%% file: sec/6_Discussion.tex
\section{Discussion and Conclusion}
\label{sec:Discussion}

Through solving blind inverse problem to recover the image under optical aberrations and motion blur, we investigate a novel and robust solution for image deblurring. The plausible performance and robustness of our proposed method are brought in by the noise schedule, which is a key component of the self-diffusion framework. We astutely observed this characteristic and boldly extended it to solving extremely unstable joint optimization inverse problems. This greatly solves the problem of unstable solutions or collased in trivial solutions. Of course, because we simultaneously constrained both the image and the kernel with neural networks and added a hierarchical noise schedule, our method requires a longer runtime compared to others. However, this time investment is worthwhile because we significantly improved the realism of the recovered image across the continuous spectrum, especially high-frequency details.

This paper presents a novel self-diffusion-based approach for optical aberration correction and motion deblurring, which we call DeblurSDI. Our method leverages the self-diffusion principle to recover accurate blur kernels and sharp images in a single framework. Experimental results on four benchmark datasets as well as six simulated optical aberrations show that DeblurSDI consistently outperforms other blind deblurring methods on various datasets, demonstrating its effectiveness and generalizability in recovering sharp images and accurate blur kernels.

%% file: main.bbl
\begin{thebibliography}{40}
\providecommand{\natexlab}[1]{#1}
\providecommand{\url}[1]{\texttt{#1}}
\expandafter\ifx\csname urlstyle\endcsname\relax
  \providecommand{\doi}[1]{doi: #1}\else
  \providecommand{\doi}{doi: \begingroup \urlstyle{rm}\Url}\fi

\bibitem[ISO(2008)]{ISO24157}
Optics and photonics -- zernike polynomials for optical surfaces and wavefronts.
\newblock ISO 24157, 2008.

\bibitem[Al~Radi et~al.(2025)Al~Radi, Majumder, and Khan]{al2025blind}
Abdul~Mohaimen Al~Radi, Prothito~Shovon Majumder, and Md~Mosaddek Khan.
\newblock Blind image deblurring with fft-relu sparsity prior.
\newblock In \emph{2025 IEEE/CVF Winter Conference on Applications of Computer Vision (WACV)}, pages 3447--3456. IEEE, 2025.

\bibitem[Aubert et~al.(2022)]{Aubert2022}
Ga{\"e}l Aubert et~al.
\newblock Product-convolution operators for microscopy.
\newblock \emph{SIAM Journal on Imaging Sciences}, 15\penalty0 (4):\penalty0 1720--1750, 2022.

\bibitem[Bigdeli et~al.(2017)]{Bigdeli2017}
Siavash~A Bigdeli et~al.
\newblock Image restoration using multi-prior gans.
\newblock \emph{CVPR}, 2017.

\bibitem[Born and Wolf(1999)]{Born1999}
Max Born and Emil Wolf.
\newblock \emph{Principles of Optics}.
\newblock Cambridge University Press, 7 edition, 1999.

\bibitem[Chen et~al.(2017)Chen, Hasinoff, et~al.]{Chen2017HDR}
Jiawen Chen, Samuel Hasinoff, et~al.
\newblock The google hdr+ pipeline.
\newblock In \emph{CVPR}, 2017.

\bibitem[Cho and Lee(2009{\natexlab{a}})]{Cho2009}
Sunghyun Cho and Seungyong Lee.
\newblock Fast motion deblurring.
\newblock In \emph{SIGGRAPH Asia}, 2009{\natexlab{a}}.

\bibitem[Cho and Lee(2009{\natexlab{b}})]{cho2009fast}
Sunghyun Cho and Seungyong Lee.
\newblock Fast motion deblurring.
\newblock \emph{ACM Trans. Graph.}, 28\penalty0 (5):\penalty0 1–8, 2009{\natexlab{b}}.

\bibitem[Chung et~al.(2023)Chung, Rombach, et~al.]{Chung2023DPS}
Hyungjin Chung, Robin Rombach, et~al.
\newblock Diffusion posterior sampling.
\newblock \emph{NeurIPS}, 2023.

\bibitem[Fergus et~al.(2006)Fergus, Singh, Hertzmann, Roweis, and Freeman]{Fergus2006}
Rob Fergus, Barun Singh, Aaron Hertzmann, Sam Roweis, and Bill Freeman.
\newblock Removing camera shake from a single photograph.
\newblock In \emph{SIGGRAPH}, 2006.

\bibitem[Gandelsman et~al.(2019)Gandelsman, Shocher, and Irani]{Gandelsman2019}
Yossi Gandelsman, Assaf Shocher, and Michal Irani.
\newblock Double-dip: Unsupervised image decomposition via coupled deep-image-priors.
\newblock In \emph{CVPR}, 2019.

\bibitem[Gonsalves(1982)]{Gonsalves1982}
R Gonsalves.
\newblock Phase retrieval and diversity in adaptive optics.
\newblock \emph{Optical Engineering}, 21\penalty0 (5):\penalty0 829--832, 1982.

\bibitem[Goodman(2005)]{Goodman2005}
Joseph~W Goodman.
\newblock \emph{Introduction to Fourier Optics}.
\newblock Roberts and Company Publishers, 2005.

\bibitem[Hasinoff et~al.(2016)Hasinoff, Durand, and Freeman]{Hasinoff2016}
Samuel Hasinoff, Fr{\'e}do Durand, and William Freeman.
\newblock Burst photography for high dynamic range and low-light imaging.
\newblock In \emph{SIGGRAPH}, 2016.

\bibitem[Hirsch et~al.(2011)Hirsch, Sch{\"o}lkopf, et~al.]{Hirsch2011}
M Hirsch, B Sch{\"o}lkopf, et~al.
\newblock Fast removal of non-uniform camera shake.
\newblock In \emph{ICCV}, 2011.

\bibitem[Hopkins(1951)]{Hopkins1951}
Harold~H Hopkins.
\newblock The diffraction theory of optical images.
\newblock \emph{Proceedings of the Royal Society of London}, 217\penalty0 (1130):\penalty0 408--432, 1951.

\bibitem[Karras et~al.(2019)Karras, Laine, and Aila]{karras2019style}
Tero Karras, Samuli Laine, and Timo Aila.
\newblock A style-based generator architecture for generative adversarial networks.
\newblock In \emph{Proceedings of the IEEE/CVF conference on computer vision and pattern recognition}, pages 4401--4410, 2019.

\bibitem[Kawar et~al.(2023)]{Kawar2023DPS}
Bahjat Kawar et~al.
\newblock Dps+: Enhanced diffusion posterior sampling.
\newblock \emph{ICML}, 2023.

\bibitem[K{\"o}hler et~al.(2012)K{\"o}hler, Hirsch, Mohler, Sch{\"o}lkopf, and Harmeling]{kohler2012recording}
Rolf K{\"o}hler, Michael Hirsch, Betty Mohler, Bernhard Sch{\"o}lkopf, and Stefan Harmeling.
\newblock Recording and playback of camera shake: Benchmarking blind deconvolution with a real-world database.
\newblock In \emph{European conference on computer vision}, pages 27--40. Springer, 2012.

\bibitem[Krishnan and Fergus(2009)]{Krishnan2009}
Dilip Krishnan and Rob Fergus.
\newblock Fast image deconvolution using hyper-laplacian priors.
\newblock In \emph{NIPS}, 2009.

\bibitem[Laroche et~al.(2024)Laroche, Almansa, and Coupete]{laroche2024fast}
Charles Laroche, Andr{\'e}s Almansa, and Eva Coupete.
\newblock Fast diffusion em: a diffusion model for blind inverse problems with application to deconvolution.
\newblock In \emph{Proceedings of the IEEE/CVF Winter Conference on Applications of Computer Vision}, pages 5271--5281, 2024.

\bibitem[Levin et~al.(2007)Levin, Fergus, Durand, and Freeman]{levin2007image}
Anat Levin, Rob Fergus, Fr{\'e}do Durand, and William~T Freeman.
\newblock Image and depth from a conventional camera with a coded aperture.
\newblock \emph{ACM transactions on graphics (TOG)}, 26\penalty0 (3):\penalty0 70--es, 2007.

\bibitem[Levin et~al.(2009)Levin, Weiss, Durand, and Freeman]{Levin2009}
Anat Levin, Yair Weiss, Fr{\'e}do Durand, and Bill Freeman.
\newblock Understanding and evaluating blind deconvolution algorithms.
\newblock In \emph{CVPR}, 2009.

\bibitem[Luo et~al.(2025)Luo, Huang, and Yang]{luo2025selfdiffusionsolvinginverseproblems}
Guanxiong Luo, Shoujin Huang, and Yanlong Yang.
\newblock Self-diffusion for solving inverse problems.
\newblock 2025.

\bibitem[Mahajan(1994)]{Mahajan1994}
Virendra~N Mahajan.
\newblock Zernike circle polynomials and optical aberrations.
\newblock \emph{Optical Engineering}, 33\penalty0 (11):\penalty0 3431--3434, 1994.

\bibitem[Mahajan(1998)]{Mahajan1998}
Virendra~N Mahajan.
\newblock \emph{Optical Imaging and Aberrations, Part I}.
\newblock SPIE Press, 1998.

\bibitem[Noll(1976)]{Noll1976}
Robert~J Noll.
\newblock Zernike polynomials and atmospheric turbulence.
\newblock \emph{Journal of the Optical Society of America}, 66\penalty0 (3):\penalty0 207--211, 1976.

\bibitem[Pan et~al.(2019)Pan, Hartley, Liu, and Dai]{pan2019phase}
Liyuan Pan, Richard Hartley, Miaomiao Liu, and Yuchao Dai.
\newblock Phase-only image based kernel estimation for single image blind deblurring.
\newblock In \emph{Proceedings of the IEEE/CVF Conference on Computer Vision and Pattern Recognition}, pages 6034--6043, 2019.

\bibitem[Paxman et~al.(1992)Paxman, Schulz, and Fienup]{Paxman1992}
Richard Paxman, Timothy Schulz, and James Fienup.
\newblock Joint estimation of object and wavefront aberrations using phase diversity.
\newblock \emph{JOSA A}, 9\penalty0 (7):\penalty0 1072--1085, 1992.

\bibitem[Platt and Shack(2001)]{PlattShack2001}
Brian Platt and Robert Shack.
\newblock History and principles of shack–hartmann wavefront sensing.
\newblock \emph{Journal of Refractive Surgery}, 17:\penalty0 S573--S577, 2001.

\bibitem[Ren et~al.(2020)Ren, Zhang, Wang, Hu, and Zuo]{ren2020neural}
Dongwei Ren, Kai Zhang, Qilong Wang, Qinghua Hu, and Wangmeng Zuo.
\newblock Neural blind deconvolution using deep priors.
\newblock In \emph{Proceedings of the IEEE/CVF conference on computer vision and pattern recognition}, pages 3341--3350, 2020.

\bibitem[Schaub et~al.(2019)Schaub, Sick, and Heintzmann]{Schaub2019}
Falko Schaub, Bernhard Sick, and Rainer Heintzmann.
\newblock Characterization of space-variant point-spread functions using eigen-psfs.
\newblock \emph{Applied Optics}, 58\penalty0 (26):\penalty0 7099--7107, 2019.

\bibitem[Shaw and Rawlins(1991)]{Shaw1991}
Peter Shaw and David Rawlins.
\newblock Three-dimensional psf measurement using microspheres.
\newblock \emph{Microscopy Research and Technique}, 19:\penalty0 410--412, 1991.

\bibitem[Smith(2007)]{Smith2007}
Warren~J Smith.
\newblock \emph{Modern Optical Engineering}.
\newblock McGraw-Hill, 4 edition, 2007.

\bibitem[Sun et~al.(2015)Sun, Cao, Xu, and Ponce]{Sun2015}
Jian Sun, Wenqi Cao, Zongben Xu, and Jean Ponce.
\newblock Learning a convolutional neural network for non-uniform motion blur removal.
\newblock In \emph{CVPR}, 2015.

\bibitem[Ulyanov et~al.(2018)Ulyanov, Vedaldi, and Lempitsky]{Ulyanov2018DIP}
Dmitry Ulyanov, Andrea Vedaldi, and Victor Lempitsky.
\newblock Deep image prior.
\newblock In \emph{CVPR}, 2018.

\bibitem[Welford(1986)]{Welford1986}
Walter~T Welford.
\newblock \emph{Aberrations of Optical Systems}.
\newblock CRC Press, 1986.

\bibitem[Whang et~al.(2022)]{Whang2022DDRM}
Jay Whang et~al.
\newblock Ddrm: Denoising diffusion restoration models.
\newblock \emph{NeurIPS}, 2022.

\bibitem[Whyte et~al.(2010)Whyte, Sivic, and Zisserman]{Whyte2010}
Oliver Whyte, Josef Sivic, and Andrew Zisserman.
\newblock Non-uniform deblurring for shaken images.
\newblock In \emph{CVPR}, 2010.

\bibitem[Wronski et~al.(2019)]{Wronski2019}
Bartosz Wronski et~al.
\newblock Handheld multi-frame super-resolution.
\newblock In \emph{SIGGRAPH}, 2019.

\end{thebibliography}
